\pdfoutput=1
\documentclass[11pt]{article}

\usepackage[]{ACL2023}

\usepackage{times}
\usepackage{latexsym}

\usepackage[T1]{fontenc}
\usepackage[utf8]{inputenc}

\usepackage{microtype}

\usepackage{inconsolata}
\usepackage{graphicx}            % 图片
\graphicspath{ {./images/} }

\usepackage[namelimits]{amsmath} %数学公式
\usepackage{amssymb}             %数学公式
\usepackage{amsfonts}            %数学字体
\usepackage{mathrsfs}            %数学花体

\usepackage{algorithm}           %伪代码
\usepackage{algorithmic}

\usepackage{multirow} % 表格
\usepackage{array}
\usepackage{booktabs}  %调整表格线与上下内容的间隔

\usepackage{float} % 固定表格位置

\title{Multi-View Graph Representation Learning for Answering Hybrid Numerical Reasoning Questions}

\author{
  Yifan Wei\textsuperscript{1,2}, Fangyu Lei\textsuperscript{1,2}, Yuanzhe Zhang\textsuperscript{1,2}, Jun Zhao\textsuperscript{1,2}, Kang Liu\textsuperscript{1,2,3}\\
  $^{1}$School of Artificial Intelligence, University of Chinese Academy of Sciences, Beijing, China\\
  $^{2}$National Laboratory of Pattern Recognition, Institute of Automation, CAS, Beijing, China\\
  $^{3}$Beijing Academy of Artificial Intelligence, Beijing, 100084, China\\
  \texttt{weiyifan2021@ia.ac.cn, leifangyu2022@ia.ac.cn} \\
  \texttt{\{yzzhang, jzhao, kliu\}@nlpr.ia.ac.cn}
}

\begin{document}
\maketitle
\begin{abstract}
Hybrid question answering~(HybridQA) over the financial report contains both textual and tabular data, and requires the model to select the appropriate evidence for the numerical reasoning task. Existing methods based on encoder-decoder framework employ a expression tree-based decoder to solve numerical reasoning problems. 
% 问题太长了
% However, the encoders rely more on Machine Reading Comprehension~(MRC) methods, which take table serialization and text splicing as input, damaging the granularity relationship between table and text as well as the spatial structure information of table itself. 
% Secondly, numerical cells and words in a large amount of hybrid data requires intricate reasoning, while table headings and other textual words are comparatively straightforward. Furthermore, fine-grained evidence such as quantities often need to be aligned with coarse-grained information like row headers. However, traditional encoders are ill-suited to handle this type of situation. 
However, encoders rely more on Machine Reading Comprehension~(MRC) methods, which take table serialization and text splicing as input, damaging the granularity relationship between table and text as well as the spatial structure information of table itself.
In order to solve these problems, the paper proposes a \textbf{M}ulti-\textbf{V}iew \textbf{G}raph (MVG) Encoder to take the relations among the granularity into account and capture the relations from multiple view.
By utilizing MVGE as a module, we constuct \textit{Tabular View}, \textit{Relation View} and \textit{Numerical View} which aim to retain the original characteristics of the hybrid data.
We validate our model on the publicly available table-text hybrid QA benchmark (TAT-QA) and outperform the state-of-the-art model.

\end{abstract}

\section{Introduction}

\begin{figure*}[t]
    \centering
    \includegraphics[width=16cm]{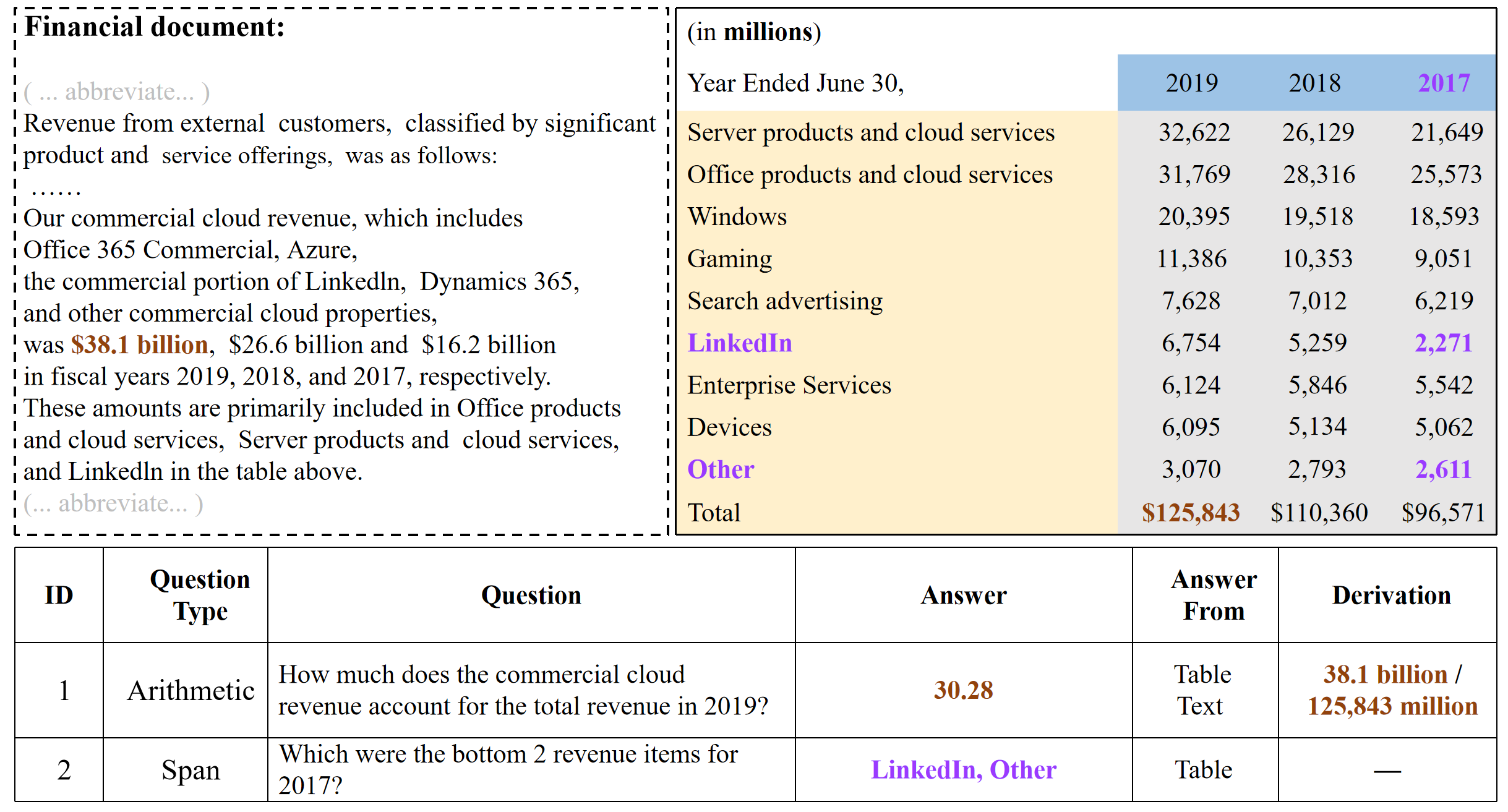}
    \caption{An example of TAT-QA. The dashed line box shows paragraphs, and the solid line box is the corresponding table.
    For the table, the row with blue background is \textit{column header} while the column with gold background is \textit{row header}. 
    The bottom table shows $Q_N$ and $Q_S$ questions that cannot be easily solved by the previous methods. The same colour marks the source of the answer.}
    \label{fig:hybrid-example}
\end{figure*}

Question Answering(QA) task aims to answer natural language questions, with evidence provided by
either plain texts \citep{rajpurkar2016squad} or structured data such as tables \citep{pasupat2015compositional,yu2018spider} or knowledge bases \citep{yih2016value,talmor2018web}.
In real applications, more QA systems need to use the heterogeneous data that combines these two types of evidence.
Thus, the hybrid form of question answering over tables and texts (HybridQA) has attracted research attention in mounting numbers \citep{chen2020hybridqa,chen2020open,chen2021finqa}. 
For example, \citet{chen2021finqa} proposed a finance benchmark FinQA, which contains questions made of many common calculations in financial analysis. 
TAT-QA \citep{zhu-etal-2021-tat} is a financial HybridQA dataset with more comprehensive compositions, Figure \ref{fig:hybrid-example} shows some examples in TAT-QA. 
%The above datasets have both textual and tabular evidences.

There are two major question types for HybridQA \citep{li2022finmath}.
One is the span selection question $Q_S$, whose answer is a span from tables and relevant texts\citep{chen2020hybridqa,chen2020open}.
The other is the numerical reasoning question $Q_N$, which usually needs to generate a numerical expression to calculate the final answer from the contents of tables and texts \citep{zhu-etal-2021-tat,chen2021finqa}.
%By contrast with $Q_S$ that has been widely studied, the $Q_N$ type question has been seldom addressed because it usually requires complex numerical reasoning and the answer cannot be obtained by simple extracting.

% only one type of data source is needed to solve them.
% Therefore, this paper mainly studies the numerical reasoning questions, especially for those complex questions across texts and tables.
% RegHNT
% \begin{figure*}[t]
%     \centering
%     \includegraphics[width=16cm]{images/reghnt-example.jpg}
%     \caption{An example of TAT-QA. The solid boxes are tables, and the dotted boxes are the corresponding paragraphs. The bottom table shows two complex questions that cannot be solved by the previous method. The same color marks the source of the answer, while the blue dashed arrow points to the source of the answer.}
%     \label{fig:hybrid-example}
% \end{figure*}

% dataset example 
% as for the question \textit{“How much does the commercial cloud revenue account for the total revenue in 2019?}, one needs to get the numerical value \textit{“125,843”} from the table, and \textit{“38.1 billion”} from the text. Then we need to generate the corresponding numerical expression \textit{“38.1 billion / 125,843 million”}. To solve such a numerical question, we need to identify the
% describing texts near the table and understand the contents of the table and the paragraphs.

% 现有工作以及他们的缺陷
% 模型方面的改变 QN-> reader QS:retriever + reader
In general, previous approaches often adopted an encoder-decoder framework. Encoders usually utilized  Transformer-based structures like TAGOP \citep{zhu-etal-2021-tat} and FinMath \citep{li2022finmath}, or GNN-based modules like GANO \citep{nararatwong2022enhancing} and RegHNT \citep{lei-etal-2022-answering} to obtain the representations of tables and texts. 
As for decoders, TAGOP \citep{zhu-etal-2021-tat} applied sequence tagging approach to extract spans. FinMath \citep{li2022finmath} used a sequence to tree framework as arithmetic decoder and a sequence tagging method as span decoder. RegHNT used a tree-based decoder to generate expressions to solve both $Q_S$ and $Q_N$.
% As to numerical reasoning questions $Q_N$, 
% % research introduces several effective methods, and these
% current approaches can be divided into three aspects\citep{wang2022survey}: encoder, decoder, and data manipulation.
%___________________________
% OUTLINES
% 当前的encoder的缺陷：
%1.…… 异构关系
%2.…… 损失表格(结构)信息
%3.…… 粒度未区分，特别是大量的数值节点representation
We note that for HybridQA, there are two main problems in the encoder part.

1) The two-dimensional tabular structure of rows and columns is damaged. Usually, the contents in tables will be flattened into sequences, and then concatenated with the texts as the encoder input \citep{zhou2022unirpg, nararatwong-etal-2022-kiqa}. As a result, the structural information of the table is lost.

2) Information of different granularity is not properly differentiated and processed. 
For example, in the case of tables in Figure \ref{fig:hybrid-example}, fine-grained evidences like purple numerical cells \textit{“2,271”, “2,611”} that can harm the graph representation do not provide meaningful knowledge unless aligned with coarse-grained attributes in row and column headers, i.e. \textit{“2017”, “LinkedIn”, “Other”}. This indicates that different granularity contributes differently according to stages.

To solve these two problems, we propose a novel method with \textbf{M}ulti-\textbf{V}iew \textbf{G}raph (\textbf{MVG}) representation learning. Multi-view graph consists of multiple views with different concerns.

% three views including \textit{Relation View}, \textit{Tabular View} and \textit{Numerical View}.
% first problem:  tabular view  ;
% second problem: relation view + numerical view（coarse-fine granularity) 
In specific, for the first problem, we design \textit{Tabular View} to enhance table structure features; 
For the second problem, we design \textit{Relation View} to help the model understand the relationship between table and text with different granularity including \{\textit{Sentence node, Row node, Column node, Number node, Word node}\}, and \textit{Numerical View} to improve the fine-grained quantity representation.

In order to integrate contributions from the three views, we utilize a multiple-view attention network \citep{xie2020mgat} to assess their contributions.
Then a sequence tagging module similar to TAGOP \citep{zhu-etal-2021-tat} is employed to obtain a span-based answer for $Q_S$.
A sequence-to-tree architecture similar to GTS \citep{xie2019goal,li2022finmath,lei2022answering} is applied in our model to generate a numerical expression tree to infer the final answers for $Q_N$. 

Experimental results on TAT-QA demonstrate that our proposed MVG model can improve EM (Exactly Match) by 20.8 and F1 values by 21.1 over the baseline, and outperforms state-of-the-art system by 0.6 and 1.1. 
% 按缺点反着描述
The main contributions of the paper are summarized as follows:
\begin{itemize}
    \item We propose a novel method with \textbf{M}ulti-\textbf{V}iew \textbf{G}raph (\textbf{MVG}) representation learning. With carefully designed view graph construction method, the proposed MVG not only keeps the two-dimensional tabular structure of rows and columns, but also properly processes different granularity information.
   
    \item We conduct extensive experiments on TAT-QA dataset, and the experimental results demonstrate MVG significantly outperforms state-of-the-art systems.
    
\end{itemize}

\section{Method}
\subsection{Preliminaries}
\textbf{Task Definition.}
Given a relevant document $D$ composed of paragraphs $P$ and numerical tables $T$, the model first aims to differentiate between the span selection question $Q_S$ and the numerical reasoning question $Q_S$.
% classify whether the question belongs to the span selection question $Q_S$ or the numerical reasoning question $Q_N$.
For $Q_S$, the model needs to select all the predicted cells from $T$ and spans from $P$ as $X = \{x_0, x_1, ..., x_n\}$.
For $Q_N$,
the model is asked to generate the answer through numerical expression $E =\{e_0, e_1, ..., e_m\}$, where $e_i$ can be a numeric value from $V^{num} \in X$, a constant quantity from $V^{con}$, or a mathematical operator from $V^{o p}=\{+, -, \times, \div, \mathrm{AVG} \}$. 

% \noindent \textbf{Overall Framework.}
% The overall architecture of MVG is given in Figure \ref{fig:framework}. Specifically, the framework consists of three model components:
% (1) a graph transformer model captures the relationship between different granularity, and integrates contributions from the multi-view graph.
% (2) a scale classifier to obtain the correct scale, and a operator classifier to categorize different types of questions.
% (3) % a dual decoder 
% a sequence tagging module to obtain answers for $Q_S$, and a tree-based decoder is applied to generate a numerical expression for $Q_N$.

\subsection{Graph Construction}
\begin{figure*}[t]
    \centering
    \includegraphics[width=16cm]{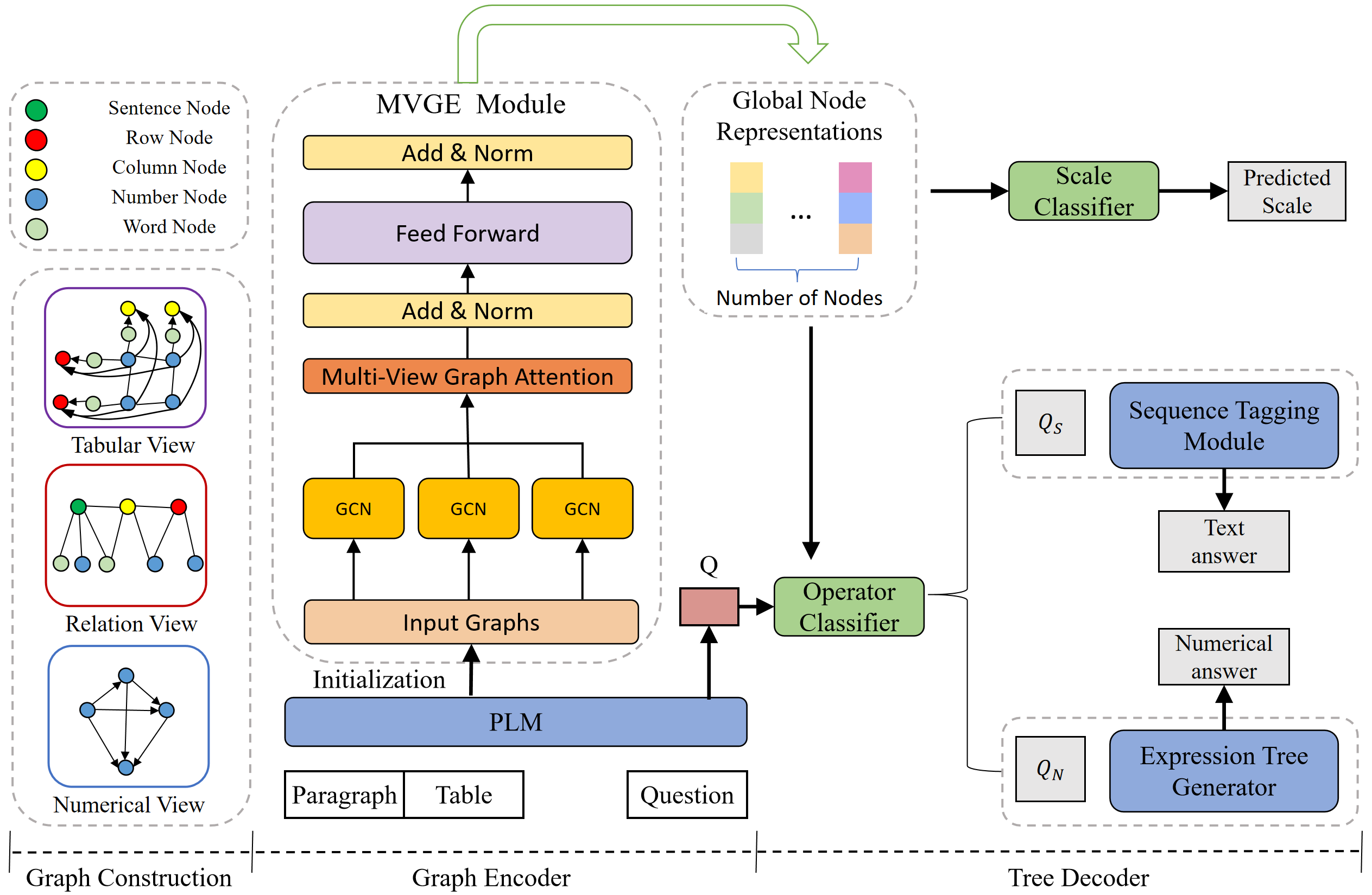}
    \caption{The overall model architecture. Our framework is divided into three parts: 1) Graph Constuction 2) Graph Encoder 3) Decoder with Sequence Tagging module and Expression Tree Generator.
    Depending on the type of question, the sequence tagging module to obtain $Q_S$ answer and tree-based decoder to get $Q_N$ answer.}
    \label{fig:framework}
\end{figure*}
Figure \ref{fig:framework} shows our proposed MVG framework. MVG first constructs multi-view graph from tables and paragraphs, which can be divided into a series of units (nodes). Each unit refers to an individual cell, a row, a column, a sentence or a word. 
% Especially, we represent the multi-view graph by K adjacency matrices,
% each adjacency matrix $ A_k \in R^{N \times N}$ is initialized with 0, where $K$ is the number of views, and $N$ is the number of nodes. And each unit can be regarded as a node $v_i \in V$ in an adjacency matrix $G = (V, E)$, and then the connection between two units becomes an edge $e = (v_i, v_j) \in E$ between their corresponding nodes. we assign value 1 to corresponding position of the adjacency
% matrix $A_{ij}$ for this edge.
Details of the connections for each view are described as the following:
\begin{itemize}
    \item \textit{Tabular-View}:
    Specially, tabular units can be divided into \textit{Cell}, \textit{Row} and \textit{Column}. And then a cell consists of words or numbers. In this paper, we connect adjacent cells as well as link a unit and its subunit, such as a column and a cell in this column. The one shows that adjacent cells are likely to describe the same attribution in row or column header. The other indicates the position of a cell in the table.
    The edge between two tabular units $U_i$ and $U_j$ is defined by:
    \begin{equation}
    A_{i j}^{tab}= \begin{cases}
    1 & \text { if $U_i$ and $U_j$ are adjacent } \\ 
    1 & \text { elif } U_i \text { is subunit of } U_j \\ 
    0 & \text { otherwise }\end{cases}
    \end{equation}
    Thus, we obtain the \textit{Tabular-View} formulated by the adjacent matrix $A^{tab}$.
    
    \item \textit{Relation-View}: 
    % 可以优化部分, Q nodes的edge设定采用<s Question>相似度，替换reghnt的 word in XXX。
    To build the relationship between table and text with different granularity, we define a \textit{Relation-View} graph and then take both coarse-grained and fine-grained characteristics into account when GNN reasoning stage.
    Specially,
    a coarse-grained node links to its fine-grained nodes, such as a sentence and words in this sentence.  And we introduce coarse-grained nodes to enrich the relationships between tables and paragraphs, which act as the intermediary that connects cells and words. For example, if a word is part of a cell, the sentence where the word is as well as the row and column which the cell is in are joined. 
    In detail,
    each granularity can be regarded as a node $v_i \in V$ in an adjacency matrix $A^{rel} = (V, E)$, and then the connection between two types of granularity becomes an edge $e = (v_i, v_j) \in E$ between their corresponding nodes. we assign value 1 to $A_{ij}^{rel}$
    Thus, we obtain the \textit{Relation-View} defined by matrix $A^{rel}$.
    % 用表来写，太复杂了
    
    % \begin{equation}
    % A_{i j}^{rel}= \begin{cases}
    % 1 & \text { if $F_j$ is subunit of $C_i$ } \\ 
    % 1 & \text { elif } U_i \text { match } U_j \\ 
    % 0 & \text { otherwise }\end{cases}
    % \end{equation}
    
    \item \textit{Numerical-View}:
    %优化，数字用ner识别和re
    The goal of the \textit{Numerical-View} is to maintain the numerical characteristics of the amount and uses heuristics \citep{zhang2020graph} to enhance representations of the relationships among fine-grained quantities. 
    We leverage a directed edge to concatenate two number nodes $N_i$ and $N_j$ as the following:
    \begin{equation}
    A_{i j}^{num}= \begin{cases}
    1 & \text { if $N_i$ > $N_j$} \\ 
    0 & \text { otherwise }\end{cases}
    \end{equation}
     Thus, we obtain the adjacent matrix, $A^{num}$, of the \textit{Numerical-View}.
    
\end{itemize}
Finally, we get \textit{Relation-View} graph, \textit{Tabular-View} graph and \textit{Numerical-View} graph.

\subsection{Node Representation Initialization}
To utilize the multi-view graph, the next process starts with initial representations of node feature matrix $X=\{x_i\}$, which are obtained using a pre-trained text encoder RoBERTa \citep{liu2019roberta} to encode each node. Specially, the size of $X$ is $N\times d$ where $d$ is defined as the dimension of input features in each node, $N$ is the total number of nodes and $i \in N $ represent the $i$-th node.
% \in R^{N \times d}$ 
% \footnote{For cell nodes, we average text embeddings of constituent cells as its node features.}.

We use the special token [SEP] as sentence node feature.
For cell nodes, we average text embeddings of constituent cells as its node features.
Especially, each cell content is a standalone short piece of text that does not necessarily benefit from contextualized text embedding by pre-trained language model (PLM) and can not reveal the hierarchical and parallel relationships between cells.
So we add row and column level nodes to solve above problems, and then initialize their representations which from pooling operations can be formulated as follows:
\begin{equation}
\mathbf{r}_n=\frac{1}{N_c} \sum_{m=1}^{N_c} \operatorname{MLP}\left({h}_{n m}\right)
\end{equation}
\begin{equation}
\mathbf{c}_m=\frac{1}{N_r} \sum_{n=1}^{N_r} \operatorname{MLP}\left({h}_{n m}\right)
\end{equation}
where $\mathbf{r}_n$ is embedding of the $n$-th row and $\mathbf{c}_m$ is embedding of the $m$-th column. $h_{nm}$ denotes the representation of each cell in the table. ${N_r}$ and ${N_c}$ represent the length of each row and the length of each column, respectively. MLP is a multi-layer perceptron with ReLU as activation fucntion.

\subsection{Multi-View Graph Encoder Module}
Once the multi-view graph is built, we first use Graph transformer to receive them. 
% In detail, this process starts with initial representations of node feature matrix $X=\{x_i\}$ by using a a pre-trained language model (PLM) such as RoBERTa \citep{liu2019roberta}. Specially, the size of $X$ is $N\times F$ where $F$ is defined as the dimension of input features in each node. 
% These encodings 
% $\{A_k\}_{k=1}^{K} 
The node feature matrix along with adjacency matrices of multi-view graph $\{A_k\}_{k=1}^{K} $ in the previous section are used as inputs to a Graph Transformer.
% Every $l$-th layer of
 The Graph Transformer incorporates a $K$-head graph convolution networks (GCNs) \citep{kipf2016semi} with multi-view graph attention layer (MGAT) \citep{xie2020mgat}, This is similar to the transformer model proposed in \citep{zhang2020graph}. Then multi-view attention networks are concatenated before a residual connection, a feedforward neural network layer (FFNN) and a layer normalization (LayerNorm) are applied.
 
In detail, we define learning of GCN  as follow:
\begin{equation}
GConv\left(A_k, X\right)=relu\left(A_k X^T W_{g k}\right)
\end{equation}
\begin{equation}
GCN\left(A_k, X\right)= {GConv}_2\left(A_k,{GConv}_1\left(A_k, X\right)\right)
\end{equation}
After that, we get the node feature matrices of the $k$-th view.
\begin{equation}
    X_{k}^{'} = GCN\left(A_k, X\right)
\end{equation}
The node representations $x_{ik}^{\prime} \in X_{k}^{'}$ are concatenated in column as $c_i$, where $i=1, 2, ..., N$, $k=1, 2, ..., K$ and $c_i$ denotes the concatenation of all view-specific representations of node $v_i$.
We defines the weight of view $k$ for node $v_i$ by applying a softmax function:
\begin{equation}
\alpha_{i k}=\frac{\exp \left(t_k^T \cdot c_i\right)}{\sum_{k^{\prime}=1}^K \exp \left(t_{k^{\prime}}^T \cdot c_i\right)}
\end{equation}
where $t_k \in  R^{d}$ is a learnable weight vector of view $k$ and $.^T$ denotes the transpose.
So far, we have obtained the weights of the nodes in different views. The final node representations can be computed by employing the attention aggregation method for multi-view.
\begin{equation}
z_{i}=\sum_{k=1}^K \alpha_{i k} \cdot x_{i k}^{\prime} \qquad   z_i \in Z
\end{equation}

At last, Graph transformer strengthens  this multi-view attention network with a feed-forward network, layer-norm layer, and residual connection:
\begin{equation}
\hat{Z}=Z+LayerNorm(Z)
\end{equation}
\begin{equation}
\bar{Z}=\hat{Z}+LayerNorm(FFN(\hat{Z}))
\end{equation}
At this point, the encoding phase is over, and then the node representations that has been enhanced can be applied for a sequence tagging module and tree-based decoder.
% $Q_S$ and $Q_N$ task.

% The graph transformer \citep{zhang-etal-2020-graph-tree} utilizes graph convolution networks (GCNs) to aggregates information about nodes and relationships of the multi-view graph.

% Each view corresponds to an independent graph convolution network, which then passes through a multi-view attention network layer.

% Each view represents a granularity relationship between nodes. Often, the granularity relationship of a single heterogeneous graph is messy, and it is easier to tell the difference with information from its fine-grained view.

% \begin{equation}
% G C N\left(A_k, X\right)=\operatorname{GConv}_2\left(A_k,  \operatorname{GConv}_1\left(A_k, X\right)\right)
% \end{equation}

% \begin{equation}
% \alpha_{i k}=\frac{\exp \left(t_k^T \cdot c_i\right)}{\sum_{k^{\prime}=1}^K \exp \left(t_{k^{\prime}}^T \cdot c_i\right)}
% \end{equation}

% \begin{equation}
% h_{i}^{\prime}=\sum_{k=1}^K \alpha_{i k} \cdot h_{i k}  
% \end{equation}

% \begin{equation}
% \hat{H}=H+\operatorname{LayerNorm}(H)
% \end{equation}

% \begin{equation}
% \bar{H}=\hat{H}+\text { Layer Norm }(F F N(\hat{H}))
% \end{equation}

%  we initializes the nodes using the pre-trained model roberta.

% In this section, we first provide a formal definition of the Numerical reasoning task for hybrid question answering.
% task
% The first line of the file must be
\subsection{Operator and Scale Classifiers}
Before the decoder generates the answer, we should classify whether the question belongs to arithmetic type or not.
TagOp defines ten operators: \textit{span-in-text, cell-in-table, spans, sum, count, average, multiplication, division, difference, and change ratio}.
In this paper, we decided to merge span-based prediction and arithmetic-based prediction. MVG outputs all predicted answer spans when it predicts the operator as \textit{span-in-text}, \textit{cell-in-table}, \textit{count} or \textit{spans} and outputs all predicted answer arithmetic when it predicts the operator as \textit{sum}, \textit{average}, \textit{multiplication}, \textit{division}, \textit{difference}, \textit{change ratio}.

For arithmetic questions, we have attained the numerical expression to generate final answer. 
However, a right prediction of a numerical answer should
not only include the right number but also the correct scale. 
Generally, the scale in TAT-QA may be \textit{None}, \textit{Thousand}, \textit{Million}, \textit{Billion}, and \textit{Percent}. 
To predict the right aggregation operator and scale, two multi-class classifiers are developed. In particular, we take the concatenated representation of [CLS], the tables and paragraphs sequentially as input to compute the probability:

\begin{equation}
    \mathrm{p}^{\mathrm{op}} =\operatorname{softmax}\left(\operatorname{FFN}\left(\left[
    \mathrm{CLS} \right]\right)\right) 
\end{equation}
\begin{equation}
    \mathrm{p}^{\text {scale }} =\operatorname{softmax}\left(\operatorname{FFN}\left(\left[\langle\mathrm{CLS}\rangle ; h_Q ; h_T ; h_P\right]\right)\right)
\end{equation}
where $h_Q$,$h_T$ and $h_P$ are the representations of the
question, the table and the paragraphs, respectively,
which are obtained by applying an average pooling
over the representations of their corresponding tokens. “;” denotes concatenation, and FFN denotes a two-layer feed-forward network with the GELU activation.

\subsection{Tree-based Decoder Module}
The tree decoder generates an equation following the pre-order traversal ordering  \citep{lei2022answering}.
In detail, for spans selection questions $Q_S$, 
we follow the baseline of the TAT-QA dataset \citep{zhu-etal-2021-tat}, the cell in the table or word in the paragraph would be regarded as answer if tokens are consecutively tagged with $I$ label.
For arithmetic questions $Q_N$, the tree decoding process involves four modules:
\begin{enumerate}
\item \textit{\underline{Attention Module}}: 
we use attention module of GTS to encode the node embedding $\bar{Z}$ to get the hybrid context state $\mathbf{{c}_{t}}$.
\begin{equation}
\begin{gathered}
\left.\alpha_{t i}=\operatorname{softmax}\left(\tanh W_{\mathrm{h}} \bar{z}_{i}+W_{\mathrm{s}}\left[\mathbf{s}_{\mathbf{t}}: \mathbf{g}_{\mathbf{t}}\right]\right)\right) \\
\mathbf{c}_{\mathbf{t}}=\sum_{i=1}^N \alpha_{t i} \bar{z}_{i}
\end{gathered}
\end{equation}

% \begin{equation} 
% \begin{split} 
%     \alpha _{ti}=\mathrm{softmax}(\mathrm{tanh}\mathrm{W_{h}}\mathbf{z_{i}}+\mathrm{W_{s}}[\mathbf{s_{t}}:\mathbf{g_{t}} ]  )) \\
%     \mathbf{c_{t}}=\sum_{i=1}^{m}\alpha _{ti}\mathbf{z_{i}} 
% \end{split}
% \end{equation}
where $W_h$, $W_s$ are weight matrices. $\alpha _{ti}$ is the
attention distribution on the node representations $\bar{z_i}$
.
\item \textit{\underline{Aggregation Module}}: 
we apply a state aggregation mechanism to calculate the expression context state $\mathbf{{g}_{t}}$.
\begin{equation}
    \mathbf{{g}_{t+1}} = \sigma({W_{g}}[\mathbf{g_{t}:g_{t,p}:g_{t,l}:g_{t,r}}])
\end{equation}
$ \sigma $ is a sigmoid function and $W_g$ is a weight matrix.
At time step 1, we use the decoder state $s_1$ to
initialize the expression context state ${g_1}$. For each node
in the currently generated expression tree, $\mathbf{g_{t,p}}$, $\mathbf{g_{t,l}}$, $\mathbf{g_{t,r}}$, represent the expression context state of the parent node, left child node, and right child
node of the current node. 
\item \textit{\underline{Decoding Module}}:
it is a bi-directional GRU, which is to generate pre-order traversal of expression trees. The hidden state $\mathbf{s_t}$ is updated as follows:
\begin{equation}
    \mathbf{{s}_{{t}+1}}=\mathrm{GRU}([\mathbf{{c}_{{t}}}:{\mathbf{g_{t}}}:\mathrm{{E}}(y_{t})],\mathbf{{s}_{{t}}})
\end{equation}
At time step 1, we use the graph representation to initialize the decoder hidden state by min-pooling operation on all node representations. Note that $\mathrm{{E}}(y_{t})$ denotes the embedding of the the token ${y_t}$.
\item \textit{\underline{Prediction Module}}: the prediction module choose to either generate a word from mathematical operators $V^{o p}=\{+,-, \times, \div, \mathrm{AVG} \}$ and a constant values set $V^{con}$, or copy a number from 
$V^{num}$ which denotes numeric values in table $T$ or paragraph $P$.
Computing a copy gate value $p_c$ to determine whether the word $y_t$ is generated or copied:
\begin{equation}
\begin{aligned}
p_c & =\sigma\left({W}_{{z}}\left[\mathbf{s}_{\mathbf{t}}: \mathbf{c}_{\mathbf{t}}: \mathbf{r}_{\mathbf{t}}\right]\right) \\
\mathrm{P}_{\mathrm{c}}\left(y_t\right) & =\sum_{y_t=\mathbf{x}_{\mathbf{i}}} \alpha_{t i} \\
\mathrm{P}_{\mathrm{g}}\left(y_t\right) & =\operatorname{softmax}\left(f\left(\left[\mathbf{s}_{\mathbf{t}}: \mathbf{c}_{\mathbf{t}}: \mathbf{r}_{\mathbf{t}}\right]\right)\right) \\
\mathrm{P}\left(y_t \mid y_{<t}, \mathrm{X}\right) & =p_c \mathrm{P}_{\mathrm{c}}\left(y_t\right)+\left(1-p_c\right) \mathrm{P}_{\mathrm{g}}\left(y_t\right)
\end{aligned}
\end{equation}
% The probability distribution
% $\mathrm{P}\left(y_t \mid y_{<t}, \mathrm{X}\right)$ of generating $y_t$ is calculated over the copy distribution $\mathrm{P}_{\mathrm{c}}\left(y_t\right)$ and generate distribution $\mathrm{P}_{\mathrm{g}}\left(y_t\right)$.

% \begin{equation}
% \begin{gathered}
% p_{c}=\sigma(\mathrm{W_{z}} [\mathbf{s_{t}:c_{t}:g_{t}} ] )\\
% \mathrm{P_{c}}(y_{t} ) =\sum_{y_{t}=\mathbf{x_{i}}}^{} \alpha _{ti}\\
% \mathrm{P_{g(y_{t})=\mathrm{softmax(f([\mathbf{s_{t}:c_{t}:g_{t}}  ])) \\
%  \mathrm{P} (y_{t}|y_{<t},\mathrm{X} )&=p_{c} \mathrm{P_{c}} (y_{t}) + (1-p_{c})\mathrm{P_{g}} (y_{t}) \\
% \end{gathered}
% \end{equation}

The final distribution is the combination of the copy probability $\mathrm{{P}_{c}}\left(.\right)$ and generated probability $\mathrm{{P}_{g}}\left(.\right)$.
\end{enumerate}
% The tree structured decoder uses the final graph layer representations $\bar{Z}$ as input and generates the target expression in $t$ time steps. At each time step $t$, let $s_t$ denote the decoding hidden state, $c_t$ denotes the hybrid context state, $g_t$ denotes the generated expressions tree state.
The algorithm for tree decoding stage like \citep{liu2019tree} is described in Algorithm~\ref{alg1}.
\begin{algorithm} 
        \renewcommand{\algorithmicrequire}{\textbf{Input:}}
	\renewcommand{\algorithmicensure}{\textbf{Output:}}
	\caption{Tree Decoding} 
	\label{alg1} 
	\begin{algorithmic}[1]
		\REQUIRE $\bar{Z},
         \mathbf{s_{t}}, \mathbf{g}_{t}$ 
		\ENSURE expression tree
        \STATE Initialize empty stack $S$
        \WHILE{S.size!=1 or S.top is not quantity }
		% \IF{$ \mathrm{p}^{\mathrm{op}} == arithmetic$} 
        \STATE $\mathbf{c_{t}},{\alpha_t}=AttentionModule(\bar{Z},\mathbf{s_{t},g_{t}})$
        \STATE{Generate $y_t$}
        \STATE ${y_t}=PredictionModule(\mathbf{s_{t},g_{t},c_{t}},{\alpha_t}) $
        \STATE $tmp = S$.top
        \STATE $S$.push($y_t$)
        \IF{$y_t \in V^{con} \cup V^{num}$}
        \WHILE{$tmp \in V^{con} \cup V^{num}$ }
        \STATE subtree $T_{sub} = S$.top[3]
        \STATE Repeat 3 Rounds: $S.$pop 
        \STATE $tmp = S$.top
        \STATE $S$.push($T_{sub}$)
        \ENDWHILE
        \ENDIF
        \STATE$ \mathbf{g_{t+1}}=Aggregation(\mathbf{g_{t},g_{t,p},g_{t,l},g_{t,r}})$
        \STATE
        $\mathbf{s_{t+1}}=Decoding(\mathbf{c_{t}},{\mathbf g_{t}},\mathrm{E}(y_{t}),\mathbf{s_{t}})$
		% \STATE $\alpha _{ti}=\mathrm{softmax}(\mathrm{tanh}\mathrm{W_{h}}\mathbf{z_{i}}+\mathrm{W_{s}}[\mathbf{s_{t}} :\mathbf{g_{t}} ]  ))$\\
  %                   $\mathbf{c_{t}}=\sum_{i=1}^{m}\alpha _{ti}\mathbf{z_{i}} $
  %           \STATE $\mathbf{\mathrm{g_{t+1}}} = \sigma(\mathrm{W_{g}}[\mathrm{g_{t}:g_{g,p}:g_{t,l}:g_{t,r}}])$
            % \IF{${ y_{t}}\in V^{op}$}
            % \STATE $\mathrm{E}(y_{t}) \gets \mathbf{M}_{op}(y_{t})$ 
            % \ELSIF{$\mathbf{\enspace y_{t}}\in V^{con}$}
            % \STATE $\mathrm{E}(y_{t}) \gets \mathbf{M}_{con}(y_{t})$ 
            % \ELSE
            % \STATE $\mathrm{E}(y_{t}) \gets h_{loc(y_{t},T,P)}^{i}$ 
            % \ENDIF
            % \STATE $\mathrm{s}_{\mathrm{t}+1}=\mathrm{BiLSTM}([\mathrm{c}_{\mathrm{t}}:g_{\mathrm {t}}:\mathrm{E}(y_{t}) ],\mathrm{s}_{\mathrm{t}})$
            % \STATE $p_{c}=\sigma(\mathrm{W_{z}} [\mathbf{s_{t}:c_{t}:g_{t}} ] )$
            % \STATE $\mathrm{P_{c}}(y_{t} ) =\sum_{y_{t}=\mathbf{x_{i}}}^{} \alpha _{ti}$
            % \STATE $\mathrm{P_{g}}(y_{t} ) = \mathrm{softmax}(f([\mathbf{s_{t}:c_{t}:g_{t}}  ]))$
            % \STATE $\mathrm{P} (y_{t}|y_{<t},\mathrm{X} )=p_{c} \mathrm{P_{c}} (y_{t}) + (1-p_{c})\mathrm{P_{g}} (y_{t}) $
            
		% \ELSE 
  % 	    \STATE Answer $\gets$ span 
		% \ENDIF 
        \ENDWHILE
        \STATE $ \mathrm{expression~tree~} T= S. $pop
		% \WHILE{$N \neq 0$} 
		% \IF{$N$ is even} 
		% \STATE $X \gets X \times X$ 
		% \STATE $N \gets N / 2$ 
		% \ELSE[$N$ is odd] \STATE $y \gets y \times X$ 
		% \STATE $N \gets N - 1$ 
		% \ENDIF 
		% \ENDWHILE 
	\end{algorithmic} 
\end{algorithm}

% \subsection{Scale Prediction}
% For arithmetic questions, we have attained the numerical expression to generate final answer. 
% However, a right prediction of a numerical answer should
% not only include the right number but also the correct scale. 
% Generally, the scale in TAT-QA may be \textit{None}, \textit{Thousand}, \textit{Million}, \textit{Billion}, and \textit{Percent}. 
% To predict the right scale, we use FNN, a two-layer feed-forward network with the GELU activation, as multi-class classifier. In particular, we take the concatenated representation of [CLS], the table and paragraphs sequentially as input to compute the probability of the scale:
% \begin{equation}
%     \mathrm{p}^{\text {scale }} =\operatorname{softmax}\left(\operatorname{FFN}\left(\left[\langle\mathrm{CLS}\rangle ; h_Q ; h_T ; h_P\right]\right)\right)
% \end{equation}
% where $h_Q$,$h_T$ and $h_P$ are the representations of the
% question, the table and the paragraphs, respectively,
% which are obtained by applying an average pooling
% over the representations of their corresponding tokens. “;” denotes concatenation.

\subsection{Training}
To optimize MVG, the overall loss is the sum of
the loss of the above three classification tasks:
\begin{equation}
\begin{aligned}
\mathcal{L} & =\mathcal{L}_{o p}+\mathcal{L}_{\text {scale }} +\mathcal{L}_{\text {tree }} \\
\mathcal{L}_{o p} & =\mathrm{NLL}\left(\log \left(\mathrm{P}^{\mathrm{op}}\right), \mathrm{G}^{\mathrm{op}}\right) \\
\mathcal{L}_{\text {scale }} & =\mathrm{NLL}\left(\log \left(\mathrm{P}^{\text {scale }}\right), \mathrm{G}^{\text {scale }}\right) \\
\mathcal{L}_{\text {tree }} & =-\sum_{t=1}^T \log \mathbf{P}\left(y_t \mid y_{<t}, \mathrm{Q}, \mathrm{T}, \mathrm{P}\right) \\
\end{aligned}
\end{equation}

where NLL(·) is the negative log-likelihood loss,
$\mathrm{G}^{\mathrm{op}}$ is from the supporting evidences which are extracted from the annotated answer and derivation. 
$\mathrm{G}^{\mathrm{scale}}$ uses the annotated scale of the answer. $\mathcal{L}_{o p}$ and $\mathcal{L}_{scale}$ are the loss functions for operator prediction and scale prediction, respectively. 
Denote the cross-entropy loss for training the tree decoder as $\mathcal{L}_{\text {tree }}$ and then the total loss can be calculated with a sum as $\mathcal{L}  =\mathcal{L}_{o p}+\mathcal{L}_{\text {scale }}+\mathcal{L}_{\text {tree }}$.

\section{Experiments}
% Our experimental setup was designed to test the effect of a multi-view graph encoder on a hybrid model.
% We choose the field of finance for our evaluation because it involves a lot of numerical information and there are still limited models for dealing with hybrid numerical documents. As for the dataset, TAT-QA provides a large sample of high quality numerical reasoning, with complex and realistic tabular and textual data.

\subsection{Dataset and Metrices}
TAT-QA \citep{zhu-etal-2021-tat} is constructed by crowd-sourcing question answer pairs on passages and tables from financial reports, which contains a total of 2,757 hybrid contexts and 16,552 corresponding question-answer pairs. And each sample contains a question, a table with 3 - 30 rows and 3 - 6 columns, and a minimum of two relevant paragraphs. 
There are four types of questions:Span, Multi-Span, Count, Arithmetic.
TAT-QA  splits into three parts, i.e., training (80\%), development (10\%), and testing (10\%).  The labels in the test set are not publicly available.
Following the previous work \citep{li2022finmath}, we utilized Exact Match (EM) and F1 score as the evaluation metrics.All of which are computed using the official evaluation script\footnote{https://github.com/NExTplusplus/TAT-QA}.On the TAT-QA challenge leaderboard\footnote{https://nextplusplus.github.io/TAT-QA/},you can view official organization rankings for test dataset performance.

% Footnotes are inserted with the \verb|\footnote| command.
% \footnote{https://nextplusplus.github.io/TAT-QA/}

\subsection{Implementation Details}
\textbf{Implementation.} 
In the MVG model, we use pre-trained language model (PLM) RoBERTa \citep{liu2019roberta} to initialize the node representations, a one layer graph transformer has a layer multi-view attentin network and 6 GCNs.
The dimensions of the hidden state for all of layers are set to 1024. 
Our model is trained for 100 epochs. The Batch size and dropout rate are set to 48 and 0.5, respectively. For optimizer, we use AdamW \citep{loshchilov2017decoupled} optimizer with a linear warmup scheduler and learning rate set to 0.0001, $\beta_1 = 0.9$ and $\beta_2 = 0.999$.
The training process of MVG model is conducted on a single RTX 3090Ti within 3 days.
During evaluation, we adopt beam search decoding with beam size 3.
\subsection{Baselines}
We compare MVG to publicly available methods as the following: 
\textbf{TAGOP}\citep{zhu-etal-2021-tat} first applies sequence tagging to extract evidences from tables and paragraphs and defines multiple operations as predictors to support discrete symbolic reasoning.
\textbf{FinMath}\citep{li2022finmath} designs a sequence-to-tree model to generate a numerical expression tree.
\textbf{KIQA} \citep{nararatwong-etal-2022-kiqa} 
utilizes pre-training model to inject external symbolic knowledge into QA model to enhance numerical reasoning.
\textbf{UniRPG} \citep{zhou2022unirpg} creates a Generator to produce executable program, and builds a executor to  perform numerical reasoning on tables and text using the programs generated in the previous phase.
\textbf{RegHNT}\citep{lei2022answering} adapts a graph to tree framwork to encode different relation types between tables and paragraphs by relation-aware attention mechanism.

\subsection{Main Results}
\begin{table}[]
\linespread{1.2} \small
%\resizebox{220pt}{30mm}{
\begin{tabular}{lcc}
\specialrule{0.1em}{0pt}{2pt}
\multirow{2}{*}{\textbf{Method}} & \multicolumn{1}{c}{\textbf{Dev}}  & \multicolumn{1}{c}{\textbf{Test}}      \\ 
% \cline{2-3} \cline{5-6} 
 & EM/F1  & EM/F1 \\  
\specialrule{0.1em}{2pt}{2pt}                       
\textbf{Baselines} &    &   \\
TAGOP \citep{zhu-etal-2021-tat}      & 55.2/62.7 &50.1/58.0 \\
FinMath  \citep{li2022finmath}          & 60.5/66.3 & 58.6/64.1 \\
KIQA   &  -/-  & 58.2/67.4 \\
GANO   &  68.4/77.8 & 62.1/71.6 \\
MHST \citep{zhu2022towards}       & 68.2/76.8 & 63.6/72.7 \\
UniPCQA \citep{deng2022pacific}  &  -/-   & 63.9/72.2 \\
UniRPG \citep{zhou2022unirpg}            & 70.2/77.9 &  67.1/76.0 \\
RegHNT \citep{lei2022answering}         & 73.6/81.3 &70.3/78.0 \\
\specialrule{0.1em}{2pt}{2pt}
MVG        & \textbf{74.5/81.5} & \textbf{70.9/79.1} \\
\specialrule{0.1em}{2pt}{0pt}
\end{tabular}
% }
\caption{\label{table-baselines}
Results of baselines and our models on the dev
and test set of TAT-QA.
}
\end{table}
% performs symbolic reasoning over the extracted evidence
% with a single type of pre-defined aggregation operators.
The evaluation results of baseline models and MVG are summarized in Table \ref{table-baselines}. Our model not only performs 20.8 and 21.1 higher on both EM and F1 compared with the original baseline (TAGOP), but also achieves the state-of-the-art results in the publicly available TAT-QA benchmark. The results demonstrates the effectiveness of MVG in numerical reasoning over tabular and textual data.

The detailed results on the test set are provided
in Table \ref{table-answer-type}. For almost all types of questions, the accuracy of MVG prediction has been improved. 
Besides, MVG model outperforms the GNN-based model GANO by a large margin, 8.8\% in terms of EM and 7.5\% in terms of F1.
Compared to RegHNT, our model performs 0.6 and 1.1 higher on both EM and F1. It demonstrates that our approach enhances the representation of coarse-grained  and fine-grained nodes.

% \usepackage{multirow}
\begin{table}[]
\resizebox{!}{32mm}{
\begin{tabular}{cccc}
\specialrule{0.1em}{0pt}{2pt}
\multirow{2}{*}{} & Table              & Text               & Table-Text         \\ \cline{2-4} 
                  & EM/F1              & EM/F1              & EM/F1      \\ \specialrule{0.05em}{2pt}{2pt}
\multicolumn{4}{c}{TAGOP}                                                        \\ \specialrule{0.05em}{2pt}{2pt}
Span              & 56.5/57.8          & 45.2/70.6          & 68.2/71.7          \\
Spans             & 66.3/77.0          & 19.0/59.1          & 63.2/76.9          \\
Counting          & 63.6/63.6          & -/-                & 62.1/62.1          \\
Arithmetic        & 41.1/41.1          & 27.3/27.3          & 46.5/46.5          \\ \specialrule{0.05em}{2pt}{2pt}
\multicolumn{4}{c}{MVG}                                                          \\ \specialrule{0.05em}{2pt}{2pt}
Span              & \textbf{76.8/78.3} & \textbf{57.0/83.7} & \textbf{81.3/86.5} \\
Spans             & \textbf{73.5/87.3} & \textbf{19.1/66.8} & \textbf{74.5/82.3} \\
Counting          & 63.6/63.6          & -/-                & \textbf{89.7/89.7} \\
Arithmetic        & \textbf{72.3/72.3} & \textbf{63.6/63.6} & \textbf{77.2/77.2} \\ \specialrule{0.1em}{2pt}{0pt}

\end{tabular}
}
\caption{\label{table-answer-type}
\small
Detailed experimental results of TAGOP and MVG w.r.t. answer types and sources on test set of
TAT-QA dataset.
}
\end{table}
% \usepackage{multirow}
% \begin{table}[H]
% \linespread{1.5} \small
% \resizebox{220pt}{!}{
% \begin{tabular}{lcccc}
% \specialrule{0.1em}{0pt}{5pt}
% \multirow{2}{*}{Answer Source} & \multicolumn{2}{c}{Spans(EM/F1)} & \multicolumn{2}{c}{Arithmetic(EM/F1)} \\
%            & TAGOP      & MVG & TAGOP     & MVG \\
% \specialrule{0.05em}{2pt}{2pt}
% Table      & 59.5/63.6  & 10.0/10.0   & 41.6/41.6 & 10.0/10.0   \\
% Text       & 43.4/69.8  & 10.0/10.0  & 27.3/27.3 & 10.0/10.0   \\
% Table-text & 66.4/73.6  & 10.0/10.0   & 48.3/48.3 & 10.0/10.0   \\
% Total      & 55.4/68.2 & 10.0/10.0   & 43.5/43.5 & 10.0/10.0   \\
% \specialrule{0.1em}{2pt}{0pt}
% \end{tabular}}
% \caption{\label{table-answer-type1}
% \small
% Detailed experimental results of TAGOP and MVG w.r.t. answer types and sources on test set of
% TAT-QA dataset.
% }

% \end{table}

\subsection{Ablation and Analysis}
To further help understand the contribution of the various components in our model, we hereby conduct ablation studies on the TAT-QA dataset. 

\noindent \textbf{Effect of Multi-View Graph Encoder.}
We first examine the contribution of relation-view Graph, tabular-view Graph and numberical-view Graph in our model. 
As expected, the performance drastically drops when ignoring the Relation-view, Tabular-view and Numerical-View from multi-view graph, leading to a relative drop of F1 by 3.9, 2.3, 1.9, respectively.
More interestingly, we also noted that enriching the quantity representation with either graph would also outperform the baseline tree-based model MHST in this task, suggesting the importance of quantity representation in hybrid numerical reasoning task.
If we delete all row and column nodes and do not use the multi-view attention, the performance drops 0.5 and 0.8 on F1. The results indicate that coarse-grained row and column nodes are important for aligning with fine-grained cells, as well as show
 different granularity contributes differently according to stages. Besides, the graph encoder is tremendously helpful to solve hybrid questions answering task.
\begin{table}[]
\begin{tabular}{lcc}
\hline 
Settings                 & EM   & F1   \\
\hline 
MVG                      & 74.5 & 81.5 \\
w/o Row and Column Nodes    & 73.8 & 81.0 \\
w/o Tabular View         & 71.9 & 79.2 \\
w/o Relation View        & 70.3 & 77.6 \\
w/o Numerical View      & 72.2 & 79.6 \\
w/o Multi-View Attention & 73.8 & 80.7 \\
\hline 
\end{tabular}
\caption{\label{ablation-study}
Ablation results on the development set of the TAT-QA dataset.
}
\end{table}

\subsection{Scale and Operater Study}
\textbf{Scale study.}
Scale prediction is a unique challenge
over TAT-QA and very pervasive in the context of
finance. After obtaining the scale, the numerical or
string prediction is multiplied or concatenated with
the corresponding scale as the final prediction to
compare with the ground-truth answer, respectively.
We compare MVG with the baseline model for
scale prediction results. The experimental results
are shown in Table \ref{scale-result}. Our model has significantly improved performance on both the dev and test
datasets. To explore the impact of the scale on results, we use the gold scale to predict the answer.
As shown in the third row of Table \ref{use-gold-label}, model accuracy will slightly increase to 84.4\% when we use
the gold scale, which shows that it is necessary to improve the prediction of scale.
\begin{table}[H]
\begin{center}
\begin{tabular}{lcccc}
\hline
Model                        &    & EM   &  & F1   \\
\hline
TAGOP                        &    & 93.5 &  &92.2 \\
RegHNT                       &    & 95.3 &  &93.4 \\
MVG                          &    & 95.4 &  &93.6 \\
\hline
\end{tabular}
\end{center}
\caption{
\label{scale-result}Scale prediction results on the dev dataset.
}
\end{table}
\noindent \textbf{Operator study.}
For TAT-QA dataset, there are four original answer types: Span, Spans, Count, Arithmetic. The paper has adapted it into two categories: Span and Arithmetic. To investigate whether this category setting causes error propagation, This paper use the gold operator to predict the answer, and the results are shown in Table \ref{use-gold-label}. When we use the gold operator, the EM and F1 of the model is improved by only 0.2 and 0.1, respectively. It suggests, to some extent, that we divide the data into two categories and use tree decoders to generate the expression for $Q_N$, and leverage sequence tagging module to solve $Q_S$ separately. This approach has no significant impact on performance.

% \subsection{Case study}

\section{Related Work}
\label{sec:bibtex}

The HybridQA is a new branch of QA task. 
As far as we know, \cite{chen2020hybridqa} proposed the first dataset of HybridQA about tables and text, and they extended it to open domain \citep{chen2020open}. They are the span selection type questions whose answer is usually a span from the table or linked paragraphs of Wikipedia. The authors designed HYBRIDER \citep{chen2020open}, a pipeline approach that divided the prediction process into two stages named linking and reasoning. 
% ___________________________________________________
\begin{table}[H]
\resizebox{!}{12mm}{
\begin{tabular}{lcc}
\hline
Model                            & EM   & F1   \\
\hline
MVG                              & 74.5 & 81.5 \\
MVG + Gold operator              & 74.7 & 81.6 \\
MVG + Gold scale                 & 77.5 & 84.4 \\
MVG + Gold operator + Gold scale & 77.6 & 84.5 \\
\hline
\end{tabular}
}
\caption{
\label{use-gold-label}The performance of using gold operators and gold scales.
}
\end{table}
Subsequently, the more tabular characteristics of numerical reasoning type questions have also been paid attention to.
TAT-QA \citep{zhu-etal-2021-tat} and FinQA \citep{chen2021finqa} builded the numerical reasoning hybrid dataset which comes from the financial field. Unlike HybridQA, they need to finish numerical calculations based on span extraction.
Both FinMath \cite{li2022finmath} and MNST \citep{zhu2022towards} had generated an expression tree explicitly to derive the final answer.
KIQA \citep{nararatwong-etal-2022-kiqa} through knowledge injection approach helpd the model to learn additional symbolic knowledge.
GANO \citep{nararatwong2022enhancing} firstly employed a GNN to integrate the structure of a table into the model’s pipeline.
RegHNT \citep{lei2022answering} focuses on designing a relation graph about the input. Besides, the graph-based encoder uses the relation-aware attention mechanism to enhance the relation representation. 
% Nevertheless, financial report is abundant with fine-grained numerical evidences, RegHNT builds nodes of different granularity into a heterogeneous graph rather than take into account the interference of these redundant numerical nodes.

\section{Conclusion}
In this paper, we propose a novel method MVG to enhance hybrid evidences representations and achieve the best performance in the publicly available TAT-QA benchmark.
We focus on improving the hybrid encoder to capture the relationship between text and table, enhancing the structural characteristics of table, and improving the representation capability of different granularity by three views. By adopting multi-view attention network, MVG could learn the importance of different views for each node, thereby distinguishing the role of different granularity according stages.
For future work, we aim to consider complex tabular layout structures such as \citep{zhao-etal-2022-multihiertt,cheng-etal-2022-hitab}  to enrich table structure features further.

\section*{Limitations}
Similar to GANO and RegHNT, our approach is to enhance the representation by constructing graph. 
And MVG has a slight improvement over RegHNT in the TAT-QA dataset, which we suspect may be due to the relational semantics in RegHNT. Therefore, MVG is more suitable as a graph encoder for node-level tasks.

Besides, the paper follows TAGOP to extract spans and then utilizes spans to generate numerical expression. 
The error from the extraction module will be transferred to the numerical inference, which will harm the accuracy of the calculation results

\bibliography{anthology,custom}
\bibliographystyle{acl_natbib}

\appendix

% \section*{Multi-view graph Detail}
% \label{sec:appendix-graph-construction}

% This is a section in the appendix.

% A coarse-grained graph: Captures the relationships between tables and paragraphs through a coarse-grained node View.
% Two fine-grained graphs:
% A numerical graph is composed of numerical nodes and their row,column and sentence nodes.
% A table structure graph consisting of the table's number nodes, word nodes, and column nodes.

% It is expected that the model will complete the interaction between text and table through {Sentence,Row,Column} three types of nodes to reduce the noise caused by redundant number nodes, where the textual attention weights decrease as the model attends more to the tabular part.
% In addition, the table retains the two-dimensional structure information of the original table through column and row node.After that, multi-view attention is used to obtain 
% the contribution of different view graph in the gnn reasoning phase.

\end{document}